\documentclass[10pt,twocolumn,letterpaper]{article}

\usepackage{iccv}              

\usepackage{algorithm}  
\usepackage{algpseudocode}

\definecolor{iccvblue}{rgb}{0.21,0.49,0.74}
\usepackage[pagebackref,breaklinks,colorlinks,allcolors=iccvblue]{hyperref}

\def\confName{ICCV}
\def\confYear{2025}

\title{Tree-NeRV: A Tree-Structured Neural Representation for Efficient Non-Uniform Video Encoding}

\author{Jiancheng Zhao \quad Yifan Zhan \quad Qingtian Zhu \quad Mingze Ma \quad 
Muyao Niu \quad Zunian Wan \\ 
Xiang Ji \quad Yinqiang Zheng\thanks{Corresponding author. Email: yqzheng@ai.u-tokyo.ac.jp}\\
The University of Tokyo, Tokyo, Japan\\
}

\begin{document}
\maketitle

\begin{abstract}
Implicit Neural Representations for Videos (NeRV) have emerged as a powerful paradigm for video representation, enabling direct mappings from frame indices to video frames. However, existing NeRV-based methods do not fully exploit temporal redundancy, as they rely on uniform sampling along the temporal axis, leading to suboptimal rate-distortion (RD) performance.
To address this limitation, we propose Tree-NeRV, a novel tree-structured feature representation for efficient and adaptive video encoding. Unlike conventional approaches, Tree-NeRV organizes feature representations within a Binary Search Tree (BST), enabling non-uniform sampling along the temporal axis. Additionally, we introduce an optimization-driven sampling strategy, dynamically allocating higher sampling density to regions with greater temporal variation. 
Extensive experiments demonstrate that Tree-NeRV achieves superior compression efficiency and reconstruction quality, outperforming prior uniform sampling-based methods. Code will be released.

\end{abstract}

\section{Introduction}
\label{sec:intro}

\begin{figure}[tbp]
  \centering
    \includegraphics[width=0.45\textwidth, height=0.18\textheight]{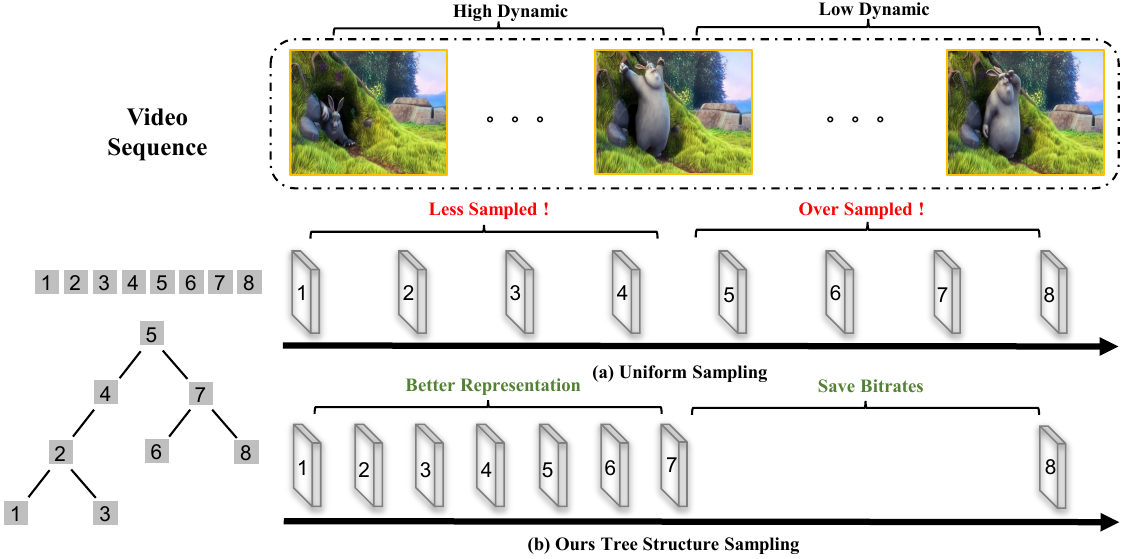}
    \caption{For a video sequence, uniform sampling tends to under-represent to the high-dynamic regions and waste bitrate in the low-dynamic ones. Whereas, our \textbf{Tree Structure Sampling} is better suited to this uneven distribution of temporal redundancies in video sequences.
    }
    \label{fig:seizure}
\end{figure}

In recent years, Implicit Neural Representation (INR) has emerged as a powerful paradigm for representing continuous signals, attracting increasing attention due to its flexibility and fast inference speed. Generally, INR models learn a continuous mapping between spatial-temporal coordinates and target values (e.g., pixel intensity, density, occupancy), typically parameterized by a multilayer perceptron (MLP).
Building on the success of INR in various vision tasks, recent studies have explored Convolutional Neural Networks (CNNs) for video representation, leading to the development of NeRV~\cite{chen2021nerv, chen2023hnerv, lee2023ffnerv, yan2024ds, li2022nerv, wu2024qs, xu2024vq, bai2023ps, zhao2024pnerv}. Unlike conventional video compression pipelines, NeRV directly learns a mapping from frame indices to video frames, leveraging CNNs to efficiently capture spational redundancy while offering simpler architectures and faster decoding speeds.
Early NeRV methods adopted Fourier-based positional encoding for frame indices. However, this approach is content-agnostic, leading to slow convergence and suboptimal performance in capturing video redunctancy.
To address these limitations, recent NeRV methods have introduced feature grids as an alternative representation~\cite{lee2023ffnerv, yan2024ds, chen2023hnerv}. When combined with CNNs, feature grids not only effectively capture spatial redundancy in video frames but also provide a more intuitive and interpretable framework by enabling direct sampling along the temporal axis.

Despite the advancements in feature grid-based NeRV models, existing methods typically structure their feature representations as \textit{linked lists}, inherently enforcing uniform sampling.
As shown in~\cref{fig:seizure}, previous methods overlook the inherently non-uniform distribution of temporal redundancy in video sequences. This strategy often results in under-sampling of high-dynamic regions while over-sampling redundant segments, leading to inefficient temporal redundancy utilization and suboptimal Rate-Distortion (RD) performance.

Therefore, in order to better align with the non-uniform temporal redundancy in video sequences, we propose Tree-NeRV, a novel tree-structured feature grid representation designed for efficient and adaptive video encoding. 
Tree-NeRV organizes and stores features within a Binary Search Tree (BST), providing a more adaptive and unconstrained representation. 
Each node in the tree consists of  a temporal key and its corresponding feature value. 
The key represents a specific time point along the video timeline, while the feature value, a three-dimensional vector, encodes spatial characteristics and serves as the basis for interpolation to compute the final time embedding. Additionally, we introduce an optimized query and balancing mechanism, ensuring efficient feature lookups, maintaining tree balance, and enabling a streamlined encoding-decoding process for improved computational efficiency.
Furthermore, to improve adaptability, we propose an optimization-driven adaptive sampling strategy, which enables Tree-NeRV to dynamically allocate higher sampling density to regions with greater temporal variation, eliminating the need for complex pre-analysis of the video sequence.
Overall, these innovations establish Tree-NeRV as a powerful solution for capturing and representing temporal variations in video sequences, achieving both high compression efficiency and superior reconstruction quality.

The main contributions of Tree-NeRV are as follows:

\begin{itemize} 
\item A tree-structured feature representation, which better aligns with the temporal characteristics of video sequence, leading to superior performance in video representation tasks compared to traditional methods. 
\item An adaptive resampling strategy during training, allowing dynamic adjustment of sampling intervals based on temporal complexity, thereby maximizing the utilization of temporal redundancy.
\item We conducted extensive experiments across multiple datasets, achieving superior performance compared to other feature grid based methods. 
\end{itemize}

\begin{figure*}[tbp]
  \centering
    \includegraphics[width=1\textwidth, height=0.32\textheight]{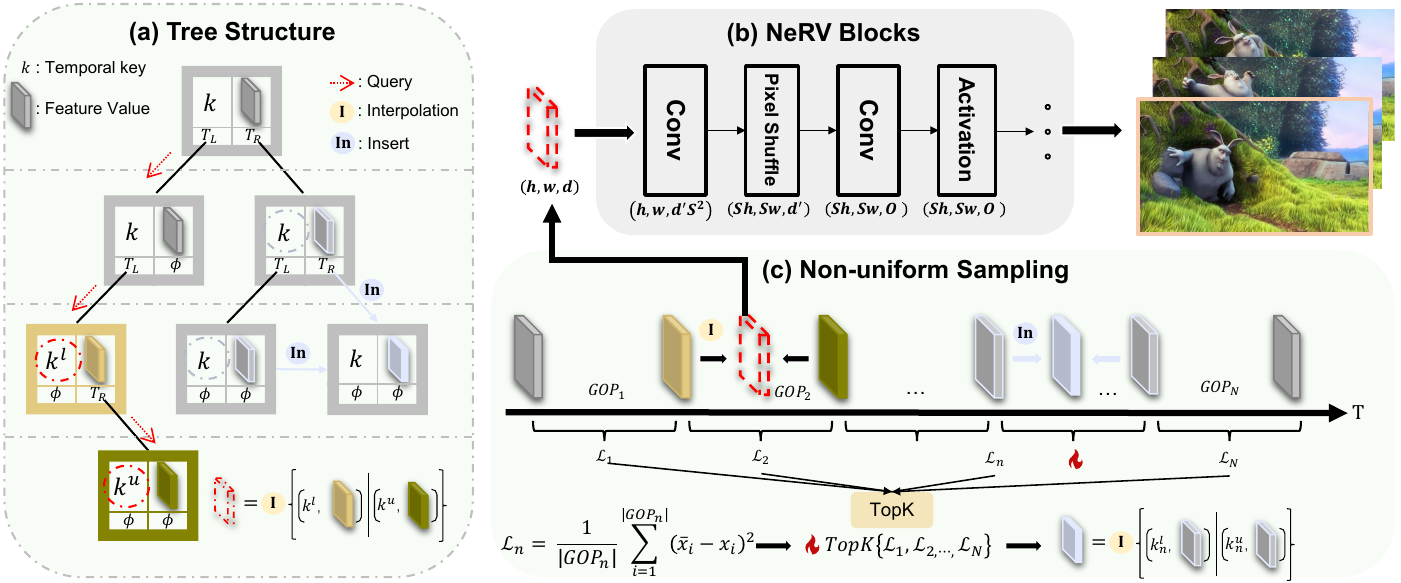}
    \caption{\textbf{Overview of Tree-NeRV.}. 
    (a) Each node in the tree $T$ structure consists of a temporal key $k$, a feature value $v$, and its left and right subtrees, $T_L$ and $T_R$, respectively. Given an query temporal index $t_i$, Tree-NeRV searches for the lower and upper bound $(k_i^l,v_i^l)$ and $(k_i^u,v_i^u)$, then performs linear interpolation between them to obtain the corresponding time embedding $v_i$ (\cref{sec:tree}); 
    (b) The time embedding $v_i$ is then processed through cascaded NeRV blocks to upsample and generate the final prediction $\hat{x}_i$ (\cref{sec:NeRV}).  
    (c) During training, an optimization-driven tree-growing and resampling strategy is employed to adaptively learn the temporal redundancy distribution of the video, allocating higher sampling density to regions with greater temporal variation (\cref{sec:tree growing}).
    .
    }
   
    \label{fig:overview}
    \vspace{-4mm}
\end{figure*}

\section{Related Work}
\label{sec:rela}

\noindent \textbf{Video Compresison.}
Video compression is a widely explored problem, encompassing well-established commercial codecs like H.264~\cite{wiegand2003overview}, H.265~\cite{sullivan2012overview}, H.266~\cite{bross2021overview}, as well as the recently flourishing data-driven codecs such as pioneering DVC~\cite{lu2019dvc} and DCVC~\cite{li2023neural}, which utilize powerful deep neural network (DNN) modules for context extraction to search for optimal coding modes and achieve extended coding capabilities~\cite{li2023neural,Sheng_2023,Li_2022}.
Although effective, both standard and data-driven codecs have shown diminishing returns in terms of rate-distortion (RD) performance as complexity increases. 
The context extraction models have become increasingly intricate and computationally inefficient, meanwhile, limiting decoding speed and yielding diminishing coding gains.

\noindent \textbf{Neural Representations for Videos (NeRV) } has achieved remarkable success in video compression, offering a simplified architecture with fast decoding speed. NeRV-like approach leverage convolutional neural networks (CNNs) to learn a temporal mapping function $f$ that directly maps frame indices to video frames while effectively capturing spatial information. Once trained, the mapping function $f$ can be parameterized as $g_{\theta}(\cdot):\mathbb{R}^t \rightarrow \mathbb{R}^{3 \times H \times W}$, where $\theta$ denotes the network parameters, $t$ corresponds to the temporal dimension, and $3 \times H \times W$ represents resolution of video frame. Furthermore, NeRV-based approaches incorporate model pruning, quantization, and entropy encoding, effectively transforming the video compression pipeline into a model compression pipeline.
Early NeRV models employ Fourier-based positional encoding as time embeddings, but this approach suffers from long training times and suboptimal convergence. To address these limitations, subsequent works have proposed various improvements: ~\citet{chen2023hnerv} introduces an content-aware autoencoder structure to capture spatial redundancy, reducing each frame to a compact latent feature representation.~\citet{zhao2023dnerv} integrates residual information between consecutive frames, improving motion representation.~\citet{lee2023ffnerv} utilizes a multi-scale resolution feature grid combined with explicit optical flow guidance to better fit video frames. The most recent~\citet{yan2024ds} achieves a weak decoupling of dynamic and static information by adopting two feature grids at different resolutions.
However, the aforementioned methods do not account for the non-uniform distribution of temporal redundancy in video sequences, which limits their ability to optimally allocate sampling resources. In contrast, Tree-NeRV is designed to better align with this characteristic, enabling adaptive and content-aware temporal sampling for more efficient video representation.

\noindent \textbf{Feature Grid Representation.}
Implicit Neural Representations (INRs) were first introduced by~\citet{mildenhall2021nerf} and have since become a widely adopted paradigm for modeling various signals, achieving notable progress in 1D audio~\cite{gao2022objectfolder20multisensoryobject}, 2D image~\cite{sitzmann2020implicitneuralrepresentationsperiodic,liu2024finer,xie2023diner} and 3D shape~\cite{park2019deepsdflearningcontinuoussigned,barron2021mip,barron2023zip,chen2022tensorf,barron2022mip}. However, these methods suffer from prohibitively long training times, limiting their practicality. To address this issue, feature grid-based representations~\cite{cao2023hexplane,fridovich2022plenoxels,fridovich2023k,muller2022instant,yu2021plenoctrees,yu2021plenoctreesrealtimerenderingneural,M_ller_2022,fridovich2023k}. have emerged as a widely adopted solution, significantly accelerating convergence speed by several orders of magnitude while maintaining high-quality reconstructions.
Recently, such representations have also been adopted for video representation tasks~\cite{chen2023hnerv,lee2023ffnerv,yan2024ds,zhao2023dnerv}, achieving remarkable success in terms of reconstruction quality and convergence speed. However, unlike 3D scene representations, video data exhibits a dense temporal-spatial distribution, where almost every pixel changing along the temporal axis, and at varying rates. Consequently, the linear interpolation commonly used in feature grid-based methods fails to effectively capture rapid temporal variations, particularly when the grid resolution is low. To overcome this limitation, Tree-NeRV introduces an adaptive sampling strategy that dynamically adjusts to the non-uniform temporal variation rates in video sequences, ensuring a more accurate and efficient representation.

\section{Method}
\label{sec:method}

\subsection{Overview}
\label{sec:overview}
In this section, we first introduce our tree-structured feature representation, including the tree architecture, and the query process for retrieving the corresponding time embedding (\cref{sec:tree}). We then discuss the NeRV block design adopted in Tree-NeRV (\cref{sec:NeRV}). Finally, we present the training strategy (\cref{sec:tree growing}), which includes the tree-growing adaptive sampling process and the balancing mechanism for maintaining tree stability when inserting new nodes during training.

\subsection{Tree Structured Feature Grid}
\label{sec:tree}

Tree-NeRV is a hierarchical representation that extends the Binary Search Tree (BST) structure. It is recursively defined as:
\begin{equation}
    T = \{ (k, v, T_L, T_R) \mid T_L, T_R \in T \}
\end{equation}
for any subtree $T$ in Tree-NeRV, it's root node contains a temporal key $k$ and a feature value $v$. The temporal key $k$ represents a specific time point in the video sequence, while the feature value $v$ is a tensor of shape $h \times w \times d$, encoding the corresponding spatial feature representation. Additionally, $T$ includes its left subtree $T_L$ and right subtree $T_R$, both of which inherit the Binary Search Tree (BST) property:
1) If the left subtree $T_L \neq \emptyset$, all nodes within $T_L$ have temporal keys strictly less than the root node's key $k$.
2) If the right subtree $T_R \neq \emptyset$, all nodes within $T_R$ have temporal keys strictly greater than the root node's key $k$.
3) Both $T_L$ and $T_R$ themselves must also satisfy the BST property.

As illustrated in~\cref{fig:overview} (a), Tree-NeRV adopts a tree-structure representation, enabling efficient temporal queries. 
Given a query time $t_i$, the objective is to locate the lower bound key $k_i^l$ and the upper bound key $k_i^u$ within the tree. These keys correspond to the feature values $v_i^l$ and $v_i^u$, respectively, which are subsequently interpolated to compute the time embedding $v_i$. 
Specifically, the search process $\mathcal{S}$ is initiated at the root node of $T$ and and proceeds recursively as follows:
\begin{equation}
\mathcal{S}(T,t_i) =
\begin{cases}
    (v_i^l, v_i^u)\leftarrow v_i,  &t_i=k, \\
    v_i^u\leftarrow v_i,  \mathcal{S}(T_L,t_i),  &t_i < k, \\
    v_i^l\leftarrow v_i, \mathcal{S}(T_R,t_i),   &t_i > k, \\
    (v_i^l, v_i^u),  &T_L = \emptyset \text{ and } T_R = \emptyset
\end{cases}
\end{equation}
1) If the temporal key $k$ of the current node matches the query time $t_i$, both bounds are set to the feature value of this node.
2) If $t_i<K$, the upper bound is updated as $v_i^u = v_i$, and the search continues in the left subtree $T_L$.
3)  If $t_i>K$, the lower bound is updated as $v_i^l = v_i$, and the search proceeds in the right subtree $T_R$.
4) The search terminates when both $T_L$ or $T_R$ reach empty, returning the current lower and upper bounds.
This recursive traversal ensures an efficient bounding mechanism, enabling fast training and inference in Tree-NeRV.
For giving a more intuitive demonstration of bound retrieval, we include pseudocode in supplementary materials~(\cref{sec:supp_treenerv}).

Once both bounds, $(k_i^l,v_i^l)$ and $(k_i^u,v_i^u)$, are determined, a linear interpolation is performed to compute the time embedding. Specifically, the relative distances between the query time $t_i$ and the two bounds are first computed as:
\begin{equation}
    d_i^l = |t_i - k_i^l|, \quad d_i^u=|k_i^u-t_i|
    \label{eq:dist}
\end{equation}
These distances serve as interpolation weights, enabling a weighted interpolation to obtain the final time embedding $v_i$:
\begin{equation}
    v_i = \frac{d_i^u}{d_i^l + d_i^u} \times v_i^l + \frac{d_i^l}{d_i^l + d_i^u} \times v_i^u
    \label{eq:interpo}
\end{equation}
The retrieved time embedding $v_i$ is subsequently processed through a series of cascaded NeRV blocks, where it is progressively upsampled to generate the final predicted video frame.

\subsection{NeRV Blocks}
\label{sec:NeRV}
We follow~\citet{li2022nerv} to design our NeRV blocks, adopting two consecutive convolutional layers with reduced channel dimensions and placing the pixel-shuffle operation in between.
The key advantage of this design is that, unlike the original NeRV block, which requires a large number of intermediate channels to support pixel-shuffle upsampling, this approach introduces an intermediate projection dimension, significantly reducing parameter overhead.
Specifically, given an input feature of shape $h\times w\times d$, 
we introduce an intermediate channel dimension of $d'S^2$, and an output channel dimension of $O$. 
Using $conv(\cdot, \cdot)$ to denote convolution kernel with corresponding input and output channel dimensions, 
our NeRV block operation is formulated as: 

\begin{equation}
    conv_{3 \times 3}(d,d'S^2) \rightarrow \text{pixel-shuffle}(S) \rightarrow conv_{3 \times 3}(d',O)
\end{equation}

The total number of trainable parameters in a single NeRV block can be computed as: 
$3 \times 3 \times d'S^2 \times (d \times s \times s + O)$, where $3 \times 3$ represents the convolutional kernel size. Compared to the parameter count of original NeRV block: $3 \times 3 \times d \times O$, 
By choosing a smaller $d'$, such as $d' = \frac{\min(d,O)}{4}$, $75\%$ reduction in parameter count can be achieved. Further discussion on NeRV are provided in the supplementary materials (\cref{sec:supp_nerv}).

\subsection{Training Tree-NeRV}
\label{sec:tree growing}
Tree-NeRV leverages a flexible and efficient tree-structured grid, enabling non-uniform sampling along the temporal axis. To further enhance adaptability while avoiding the need for explicit time complexity analysis of a given video sequence, we introduce an optimization-driven Tree Growing and Resampling strategy. This strategy follow a coarse to fine principle, allows Tree-NeRV to dynamically adjust its sampling strategy based on the training process.

\noindent\textbf{Warm-up Stage:}
Given a video sequence of length $L$, denoted as $\mathbb{V} = \{ x_i \}^{L-1}_{i=0}$, where each frame $x_i \in \mathbb{R}^{ 3 \times H \times W}$,
we first initialize the tree with a coarse and uniform sampling interval of $\frac{L}{N}$, where $N\ll L$. 
By treating each pair of adjacent features as natural boundaries, this process partitions $\mathbb{V}$ into $N$ Groups of Pictures (GOPs):
\begin{equation} 
{ \overbrace{x_0, \dots, x_{ \frac{L}{N}-1}}^{\text{GOP}_1} \mid \overbrace{x_{\frac{L}{N}}, \dots, x_{2\frac{L}{N}-1}}^{\text{GOP}_2} \mid \dots \mid \overbrace{x_{L-\frac{L}{N}}, \dots, x_{L-1}}^{\text{GOP}_N} }
\end{equation}
During this warm-up stage, the initial structure provides a coarse representation of the video sequence.

\noindent\textbf{Tree-growing Stage:}
After the warm-up phase, the Tree Growing stage is activated. At this stage, Tree-NeRV adaptively refines the sampling strategy by allocating more bitrate to regions that are underrepresented during reconstruction. This allocation is guided by the reconstruction error computed during training:
\begin{equation}
    \mathcal{L}_n  = \frac{1}{|GOP_n|} \sum_{i=0}^{|GOP_n|} (x_i - \hat{x}_i)^2.
\end{equation}
where $n$ denotes the $n$-th $GOP$, $|\cdot|$ denotes the frames included in $GOP_n$, and $\mathcal{L}_n$ represents the average reconstruction error of frames within $GOP_n$. High-error GOPs are identified as highly dynamic regions in the video sequence, where the initial sampling density may be insufficient. To address this, we select the $TopK$ GOPs with the highest reconstruction error and perform denser sampling in those regions:
\begin{equation}
    TopK\{ \mathcal{L}_1, \mathcal{L}_2, \ldots, \mathcal{L}_N\}
\end{equation}

For each selected high-error $GOP_n$, its temporal boundaries are represented as $(k_{n}^l,v_{n}^l)$ and $(k_{n}^u,v_{n}^u)$. Then a new node is inserted between $k_n^l$ and $k_n^u$. In our experiments, we choose the midpoint of the interval:
\begin{equation} 
k_{n}^{\text{In}} = \frac{k_n^u + k_n^l}{2} 
\end{equation}
The corresponding feature value $v_n^\text{In}$ for the new node is obtained via linear interpolation, following~\cref{eq:dist} and~\cref{eq:interpo}. This ensures that the tree structure is refined in a temporally adaptive manner, improving reconstruction fidelity in high-dynamic regions of the video.

\noindent\textbf{Tree Structure Balance:}
After insert a new node, Tree-NeRV may become unbalanced, potentially degrading query efficiency. To address this, we introduce the AVL~\cite{adel1962algorithm} tree balancing mechanism, ensuring that the tree remains well-structured and efficient for traversal. For a more in-depth discussion of the AVL balancing mechanism in Tree-NeRV, we provide a detailed analysis in the supplementary materials~(\cref{sec:supp_balance}). Here, we present two illustrative examples to demonstrate the rebalancing process.

\begin{figure}[tbp]
  \centering
    \includegraphics[width=0.45\textwidth, height=0.09\textheight]{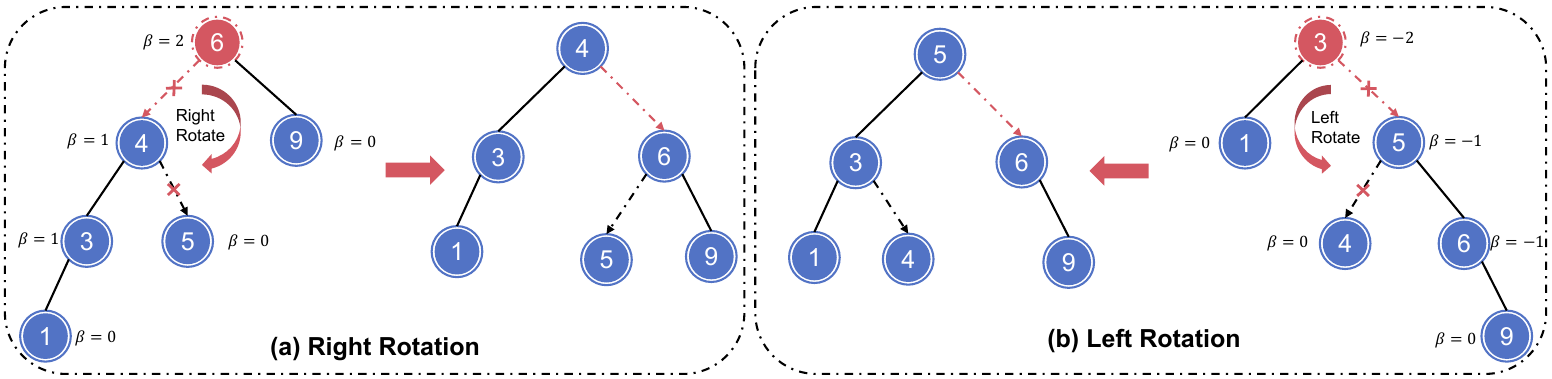}
    \caption{(a) and (b) illustrate two types of rotation operations used for rebalancing. For simplicity, nodes are represented by their keys. \textcolor{red}{Red nodes} indicate unbalanced nodes, while \textcolor{blue}{blue nodes} represent balanced nodes. Dashed lines depict connections that have been modified during the rotation process.
    }
    \label{fig:rotate}
    \vspace{-3mm}
\end{figure}
To maintain the balance of the tree structure, we introduce two types of rotation operations, as illustrated in \cref{fig:rotate}. \textsc{Right Rotation} (~\cref{fig:rotate} (a)): the tree becomes left-heavy, with node $6$ identified as the unbalanced node. To restore balance, a right rotation is performed at node $6$, repositioning it as the right child of node $4$, Simultaneously, the original right child of node $4$ is reassigned as the left child of node $6$ after rotation. 
\textsc{Left Rotation} (~\cref{fig:rotate} (b)): the tree becomes right-heavy, with node $3$ as the unbalanced node. To rebalance, a left rotation is applied at node $3$, making it the left child of node $5$. Meanwhile, the original left child of node $5$ is reassigned as the right child of node $3$ after rotation.

\section{Experiment}
\label{sec:exp}

In this section, we present the experimental setup for reproducibility, including datasets, procedures, and hyperparameter configurations (\cref{sec:setup}). We then present the video representation results with both quantitative metrics and visualizations (\cref{sec:representation}). Next,we analyze the correlation between Tree-NeRV’s sampling patterns and temporal variations (\cref{sec:sampling}),  followed by an evaluation of its encoding and decoding efficiency (\cref{sec:decoding}). 
Additionally, we assess Tree-NeRV’s performance on downstream tasks, including video interpolation (\cref{sec:interpo}) and video compression (\cref{sec:compression}), Finally, we conduct an ablation study on the effectiveness of our adaptive sampling (\cref{sec:abla}).

\subsection{Setup}
\label{sec:setup}

\begin{table*}[!t]
    \begin{subtable}{.5\linewidth}
    \centering
    \tabcolsep=0.22cm
    \resizebox{!}{1.2cm}{
            \begin{tabular}{c|c c c c c } \hline 
            sizes& 0.35M& 0.75M& 1.5M&3M \\ \hline 
            NeRV~\cite{chen2021nerv}& 26.59& 28.70& 30.60&34.37\\ 
            FFNeRV~\cite{lee2023ffnerv}& 28.08 & 31.01& 33.96&36.85\\
             HNeRV~\cite{chen2023hnerv}& 29.20& 32.38& 33.68&36.59\\
            DS-NeRV~\cite{yan2024ds}& 29.78& 32.35& 35.03&36.85\\ \hline 
            \textbf{Ours}& \textbf{30.21}& \textbf{33.14}& \textbf{36.21}&\textbf{38.68}\\ \hline
            \end{tabular}
    }
    \caption{PSNR($\uparrow$) on Bunny (720P) with varying \textbf{model size}.}
    \label{tab:bunny}
    \end{subtable}%
    \begin{subtable}{.5\linewidth}
    \centering
    \tabcolsep=0.22cm
    \resizebox{!}{1.2cm}{
            \begin{tabular}{c|c c c c c}
            \hline
                epochs & 100 & 150 & 200 & 250 & 300  \\ \hline
                NeRV~\cite{chen2021nerv} & 24.89 & 25.72 & 26.26 & 26.53 & 26.59  \\ 
                FFNeRV~\cite{lee2023ffnerv} & 26.62 & 27.44 & 27.86 & 28.02 & 28.08  \\ 
                HNeRV~\cite{chen2023hnerv} & 27.07 & 28.26 & 28.99 & 29.12 & 29.20  \\
                DS-NeRV~\cite{yan2024ds} & 28.48 & 29.14 & 29.44 & 29.69 & 29.78 \\ \hline
                Ours & \textbf{28.78} & \textbf{29.48} & \textbf{29.88} & \textbf{30.13} & \textbf{30.21}  \\ \hline
            \end{tabular}
    }
    \caption{PSNR($\uparrow$) On Bunny (720P) with varying \textbf{epochs}.}
    \label{tab:epochs}
    \end{subtable}%
    \caption{Video \textbf{reconstruction} results on Bunny.}
\end{table*}

\begin{table*}[!t]
    \centering
    \tabcolsep=0.22cm
    \begin{tabular}{c|c c c c c c c| c}
            \hline
            Video & Beauty & Bosph & Honey & Jockey & Ready & Shake & Yacht & avg.\\ 
            \hline 
            NeRV~\cite{chen2021nerv} & 32.79 & 31.98 & 37.91 & 30.04 & 23.48 & 32.89 & 26.26 & 30.76\\ 
            FFNeRV~\cite{lee2023ffnerv} & 33.37 & 35.03 & 38.95 & 32.22 & 26.58 & 33.82 & 28.62 & 32.66\\
            HNeRV~\cite{chen2023hnerv} & 31.37 & 35.03 & 38.20 & 31.58 & 25.45 & 34.89 & 28.98 & 32.21\\
            DS-NeRV~\cite{yan2024ds} & 33.29 & 34.31 & 38.98 & 32.65 & 26.41 & 34.04 & 28.72 & 32.63\\ 
            \hline 
            \textbf{Ours} & \textbf{33.54}& \textbf{35.63}& \textbf{39.88}& \textbf{32.74}& \textbf{26.86}& \textbf{35.28}& \textbf{29.74}& \textbf{33.36}\\
            \hline
            \end{tabular}
    \caption{Video \textbf{reconstruction} results on UVG (1080P), PSNR($\uparrow$) reported.}
    \label{tab:uvg_recons}
    \vspace{-3mm}
\end{table*}

\begin{table*}[!t]
    \centering
    \tabcolsep=0.22cm
    \begin{tabular}{c|c c c c c c c c c c c}
    \hline
        Video & b-swan & b-trees & boat & b-dance & camel & c-round & c-shadow & cows & dance & dog & avg.  \\ \hline
        NeRV~\cite{chen2021nerv} & 25.04 & 25.22 & 30.25 & 25.78 & 23.69 & 24.08 & 25.29 & 22.44 & 25.61 & 27.15 & 25.30  \\ 
        HNeRV~\cite{chen2023hnerv} & 26.42 & 26.96 & 27.60 & 31.90 & 26.24 & 27.18 & 27.55 & 25.27 & 28.48 & 27.60 & 27.52  \\ 
        FFNeRV~\cite{lee2023ffnerv} & 31.24 & 28.73 & 33.52 & 32.18 & 25.74 & 28.50 & 33.88 & 24.14 & 28.42 & 30.64 & 29.70  \\ 
        DS-NeRV~\cite{yan2024ds} & 31.55 & 29.96 & 33.18 & 32.72 & 26.48 & 29.23 & 35.33 & 24.59 & 28.17 & 32.54 & 30.38  \\ \hline
        Ours & \textbf{32.72} & \textbf{31.55} & \textbf{34.64} & \textbf{32.74} & \textbf{27.90} & \textbf{30.71} & \textbf{36.32} & \textbf{26.09} & \textbf{29.75} & \textbf{34.02} & \textbf{31.64} \\ 
    \end{tabular}
    \caption{Video \textbf{reconstruction} results on DAVIS (1080P), PSNR($\uparrow$) reported.}
    \label{tab:DAVIS_recons}
    \vspace{-3mm}
\end{table*}

\textbf{Datasets.} 
To evaluate the effectiveness of the proposed Tree-NeRV, we conduct both quantitative and qualitative comparison experiments on the Big Buck Bunny~\cite{bigbuckbunny}, UVG~\cite{mercat2020uvg}, and DAVIS~\cite{wang2016mcl} datasets. Big Buck Bunny consists of 132 frames with a resolution of $720 \times 1280$. UVG contains seven videos with a resolution of $1080 \times 1920$, each with a duration of 5 or 2.5 seconds at 120 fps. Additionally, following~\cite{yan2024ds}, we selected 10 videos from the DAVIS validation subset, each with a resolution of $1080 \times 1920$ and frame counts ranging from 50 to 200 frames, for additional experiments.

\noindent\textbf{Implementation Details.} 
We evaluate video representation quality using Peak Signal-to-Noise Ratio (PSNR), while bits per pixel (bpp) is used as an indicator of video compression performance. For optimization, we utilize the Adan~\cite{xie2024adan} optimizer with L2 loss to measure the pixel-wise difference between predictions and ground truth. The learning rate is initialized to $1 \times 10^{-2}$ and follows a cosine annealing schedule, with a batch size of 1 applied across all datasets. Unless otherwise noted, Tree-NeRV is initialized with a uniform coarse sampling ratio $0.1\times L$, with a growth interval of 10 epochs applied consistently across all datasets. All models have 3 million (3M) parameters and are trained for 300 epochs. Experiments are conducted on a single NVIDIA GTX 4090.

\subsection{Video Representation}
\label{sec:representation}
As shown~\cref{tab:bunny}, we first compared our proposed Tree-NeRV with NeRV~\cite{chen2021nerv}, FFNeRV~\cite{lee2023ffnerv}, HNeRV~\cite{chen2023hnerv}, and DS-NeRV~\cite{yan2024ds} on the video representation task, covering both positional encoding-based and feature grid-based approaches. On the Big Buck Bunny dataset, we evaluated the models across different sizes, ranging from 0.3M to 3M parameters, to assess their reconstruction performance. Additionally, we compared model convergence at different epochs using a fixed parameter size of 0.3M. The results in~\cref{tab:epochs} demonstrate that Tree-NeRV outperforms other methods consistently across all model sizes, and exhibits significantly faster convergence at varying epochs.

\begin{figure*}
    \centering
    \includegraphics[width=0.98\textwidth]{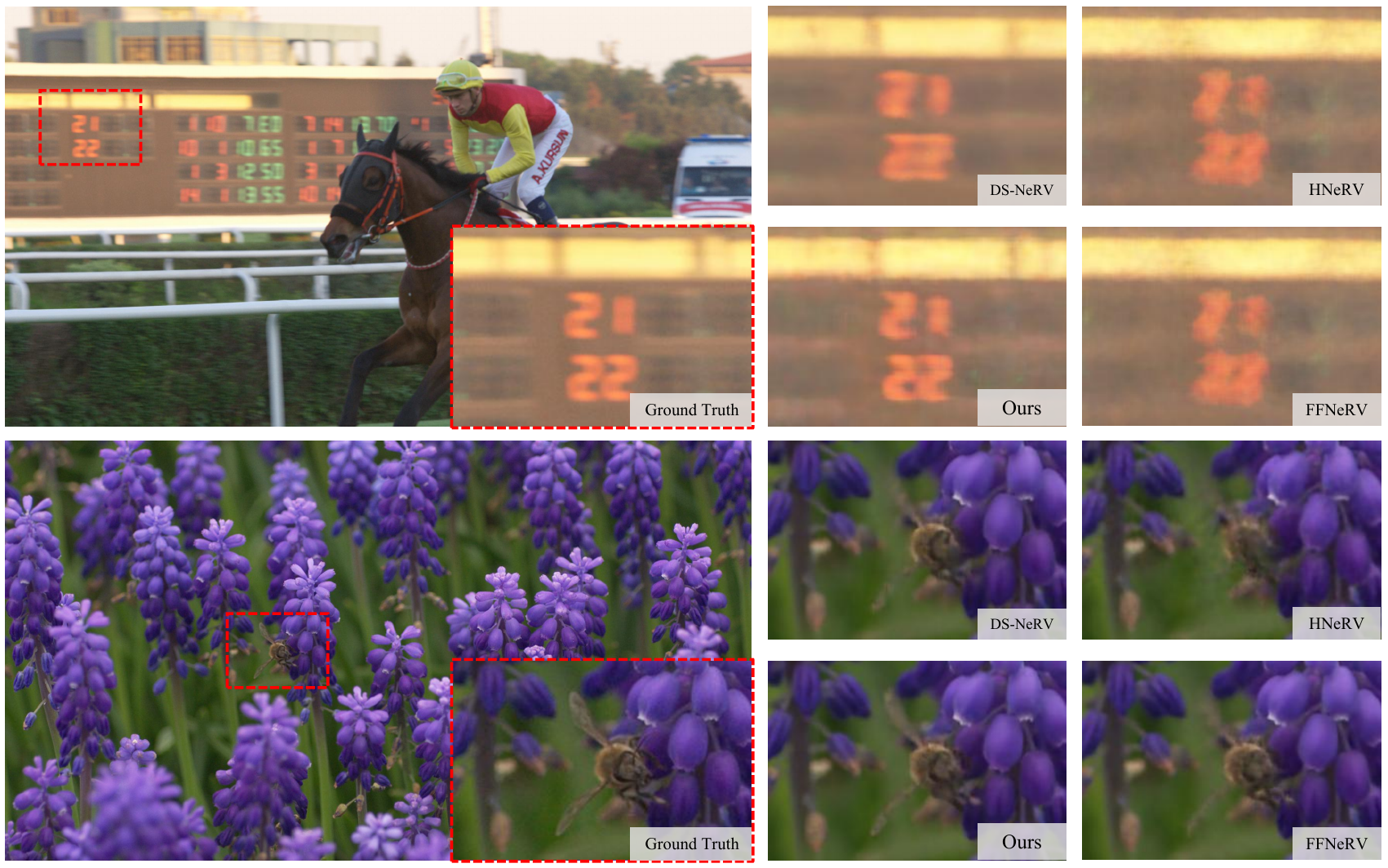}
\caption{\textbf{Video representation results.} Other methods failed to capture certain details, such as the digits on the scoreboard in `Jockey' (Top) and the intricate wing structure of the honeybee in `Honeybee' (Bottom). In contrast, our method effectively captured and reconstructed these fine details.
    }
  \label{fig:recon}
\end{figure*}

To further validate our approach, we extend our evaluation to the UVG~\cite{mercat2020uvg} and DAVIS~\cite{wang2016mcl} datasets. The quantitative results are summarized in~\cref{tab:uvg_recons} and~\cref{tab:DAVIS_recons}. On average, Tree-NeRV achieves a PSNR improvement of 0.73dB over competing methods on UVG and 1.26dB on DAVIS, demonstrating its superior reconstruction quality. 
We further visualize two representative cases in~\cref{fig:recon}: `Jockey', which has rapid camera movement, and `Honeybee', where the camera remains static. In these cases, Tree-NeRV effectively reconstructs fine details, such as the scoreboard digits in `Jockey' and the intricate wing structures in `Honeybee', where other methods struggle with visible artifacts and fail to achieve comparable reconstruction quality.
We also provide additional visual comparisons in the supplementary material~{\cref{sec:supp_Qualitative result}}. Please refer to it for more details.

\begin{figure*}
    \centering
    \includegraphics[width=0.95\textwidth]{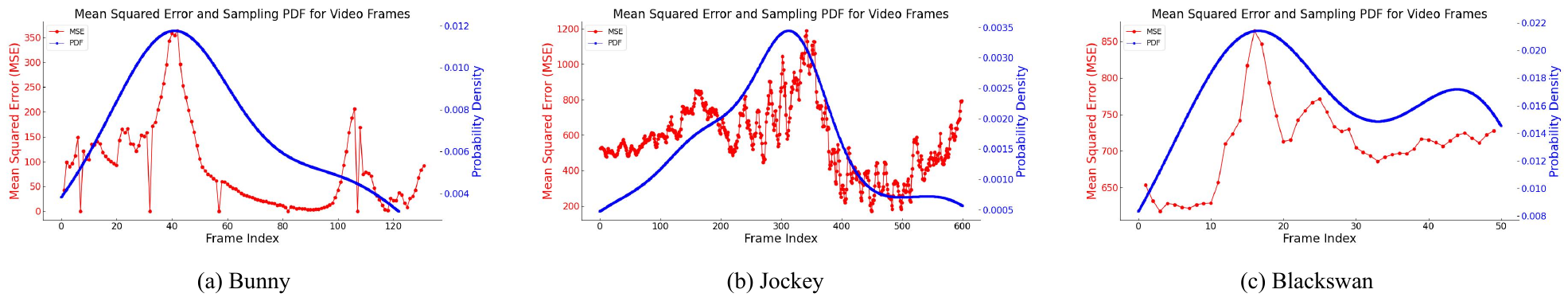}
    \caption{\textbf{Sampling results of our Tree-NeRV.} In the figure, the \textcolor{red}{red line} represents the temporal variation of the video sequence, quantified by the mean squared error (MSE) between adjacent frames. The \textcolor{blue}{blue line} depicts the probability density function (PDF) of Tree-NeRV's actual sampling points. It is evident that Tree-NeRV's sampling density aligns closely with the temporal variation trends of the video sequence.
    }
  \label{fig:sampling}
\end{figure*}

\begin{figure*}
    \centering
    \includegraphics[width=0.95\textwidth]{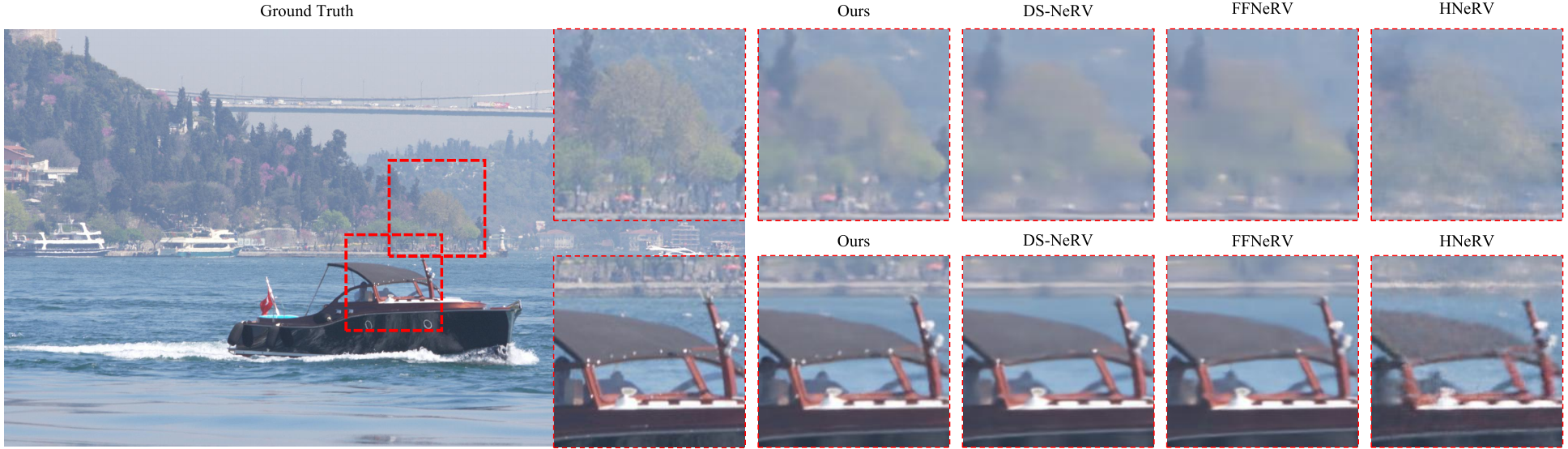}
    \caption{\textbf{Video interpolation results on `Bosphorus', interpolated frame shown above.} Compared with other methods, our approach successfully reconstructs the fine details of the boat's top and the background in the interpolated frame, whereas other methods exhibit more severe artifacts and suffer significant detail loss.
    }
  \label{fig:interpolation}
\end{figure*}

\begin{table*}[t]
    \centering
    \tabcolsep=0.22cm
        \begin{tabular}{c|ccccccc|c}
        \hline
            video & Beauty & Bosph & Honey & Jockey & Ready & Shake & Yacht & avg. \\ \hline
            FFNeRV~\cite{lee2023ffnerv} & 31.14&32.66&38.15&21.98 & 19.44& 29.39& 26.49& 28.46\\ 
            HNeRV~\cite{chen2023hnerv} & 31.16& 31.72& 38.10& \textbf{23.82}& \textbf{22.39}& 32.34& 27.26&29.57\\ 
            DS-NeRV~\cite{yan2024ds} & 31.43& 33.89& 38.69&21.74&20.57& 32.17& 27.14& 29.38\\ \hline
            Ours & \textbf{31.99}&\textbf{35.44}& \textbf{39.84}& 22.24& 20.73&\textbf{32.78}&\textbf{28.10}& \textbf{30.16}\\ \hline
        \end{tabular}
    \caption{Video \textbf{interpolation} results on UVG, PSNR($\uparrow$) reported.}
    \label{tab:interpolation}
    \vspace{-3mm}
\end{table*}

\subsection{Sampling Results}
\label{sec:sampling}

To intuitively visualize the sampling behavior of Tree-NeRV, we conducted experiments to analyze its sampling distribution. First, we quantified temporal variations in the video sequence using frame-wise residuals. Specifically, we computed the residuals between adjacent frames and measured their magnitude using Mean Squared Error (MSE) as a metric.
We then plotted the MSE variations in~\cref{fig:sampling}, where the horizontal axis represents time, and the vertical axis shows the MSE, with the red line highlighting the variation.

After training, we further analyzed the sampling distribution of Tree-NeRV and plotted the probability density function (PDF) of the temporal sampling points, as represented by the blue line in~\cref{fig:sampling}. It can be observed that the temporal trend of video residuals closely aligns with the sampling density of Tree-NeRV, where regions with higher temporal variations correspond to a higher sampling density.

\subsection{Video Encoding and Decoding}
\label{sec:decoding}
To evaluate the efficiency of the proposed tree-structured feature grid representation, we compared Tree-NeRV against other methods. 
The comparison was conducted in terms of both encoding and decoding performance. All models were trained for 300 epochs with a batch size of 1 using a single Nvdia GTX 4090 GPU, and we measured both the total training time and the decoding frame rate (FPS) of the final trained model.

As summarized in~\cref{tab:endecoding}, Tree-NeRV demonstrated superior encoding and decoding efficiency compared to other approaches. HNeRV, by storing a unique latent feature for each time step, achieved the highest decoding speed, as it bypassed the need for querying and interpolation during inference. However, this came at the cost of significantly increased training time, due to the reliance on an additional encoder to extract latent features from each input frame. 
Additionally, We also provide an evaluation of Tree-NeRV’s convergence performance in terms of the achieved quality within a fixed encoding time, as presented in~\cref{tab:enc_time}. Notably, our method reaches 30dB PSNR within just 10 minutes of training.

\begin{table}[!t]
    \centering
    \tabcolsep=0.15cm
    \renewcommand{\arraystretch}{1.2} 
    \begin{tabular}{c|c|c}
        \hline
        Time & Encoding Time (h)($\downarrow$) & Decoding FPS($\uparrow$) \\ 
        \hline 
        FFNeRV~\cite{lee2023ffnerv} & $\thicksim6h$ & 54.68 \\
        HNeRV~\cite{chen2023hnerv} & $\thicksim 6h$ & \textbf{105.90} \\
        DS-NeRV~\cite{yan2024ds} & $\thicksim 3h$ & 59.33 \\ 
        \hline 
        \textbf{Ours} & $\thicksim \textbf{2h}$ & 70.14 \\ 
        \hline
    \end{tabular}
    \caption{Video \textbf{encoding} and \textbf{decoding} speed comparison on UVG. Encoding time (h) and decoding FPS are reported.}
    \label{tab:endecoding}
    \vspace{-3mm}
\end{table}

\begin{table}[!t]
    \centering
    \renewcommand{\arraystretch}{1.2} 
    
    \begin{tabular}{l|cccc}
        \hline
        Encoding Time & 5 min & 10 min & 1 hour & 2 hours \\ 
        \hline 
        Tree-NeRV (Ours) & 28.00 & 30.11 & 32.54 & 33.36  \\
        \hline
    \end{tabular}

    \captionsetup{font=scriptsize}
    \caption{Tree-NeRV performance on UVG dataset at different encoding times.}
    \label{tab:enc_time}
    \vspace{-3mm}
\end{table}

\subsection{Video Interpolation}
\label{sec:interpo}
For downstream tasks, we conducted video interpolation experiments on the UVG dataset, following the settings of previous works~\cite{yan2024ds, lu2022video, zhang2023extracting}. Specifically, we trained the model on even-numbered frames and tested on odd-numbered frames. To ensure a fair comparison, we applied direct interpolation on the grid for HNeRV, avoiding the use of frames as inputs.
We provide both quantitative and qualitative evaluations. As shown in~\cref{tab:interpolation}, Tree-NeRV outperforms DS-NeRV with a 0.78 dB improvement in PSNR. Visual results in~\cref{fig:interpolation} demonstrate that Tree-NeRV effectively reconstructs intricate details on the top of the boat and captures complex variations in the background, yielding superior interpolation quality compared to other methods.

\subsection{Video Compression}
\label{sec:compression}
\begin{figure}
    \centering
\includegraphics[width=0.5\textwidth]{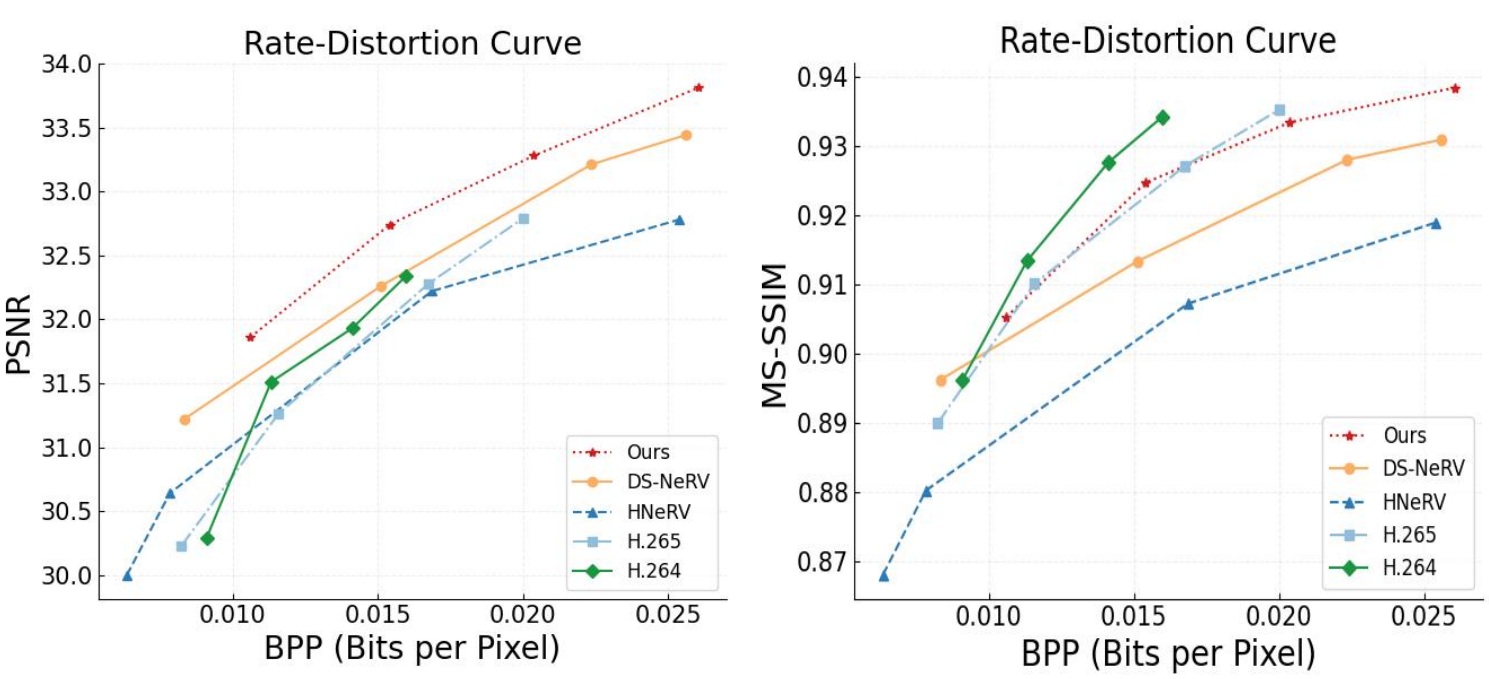}
    \caption{R-D Curve on UVG dataset.}
  \label{fig:RD}
  \vspace{-3mm}
\end{figure}

For video compression, we followed the standard pipeline~\cite{chen2021nerv} consisting of `model pruning (0.1) - quantization (8 bits) - entropy coding' pipeline. We compared the performance of Tree-NeRV against traditional codecs (H.264~\cite{wiegand2003overview} and H.265~\cite{sullivan2012overview}) and INR-based methods (DS-NeRV and HNeRV). As depicted in~\cref{fig:RD}, the bitrate per pixel (bpp) curves demonstrate that Tree-NeRV not only surpasses other INR-based approaches but also achieves rate-distortion (RD) performance that is comparable to the standard codecs.

\subsection{Ablation Studies}
\label{sec:abla}
In this section, we compared Tree-NeRV with uniform sampling of varing sampling quantities. As shown in~\cref{tab:abla}, Tree-NeRV, using only 100 sampling points, outperform the performance of a uniformly sampled method with $1.5 \times$ more sampling points and achieves comparable results to $2 \times$ sampling points. These findings further demonstrate the effectiveness of our proposed adaptive sampling strategy in enhancing model efficiency and compression performance.

\begin{table}[!t]
    \centering
    \tabcolsep=0.15cm
    \renewcommand{\arraystretch}{1.2} 
    \begin{tabular}{c|ccc|c}
        \hline
        \#Feature & 100 & 150 & 200 & Ours\\ 
        \hline 
        Model Size & 2.99M & 3.71M & 4.43M & 2.99M\\
        \hline
        PSNR & 32.11 & 33.33 & \textbf{33.42} & 33.36\\
        \hline
    \end{tabular}
    \caption{Comparison of Our Method with Uniform Sampling at Different Feature Quantities on UVG. Ours sampled 100 feature points after training.}
    \label{tab:abla}
    \vspace{-3mm}
\end{table}

\section{Conclusion}
\label{sec:conclu}
In this paper, we introduce Tree-NeRV, a more flexible and unconstrained tree-structured feature grid representation for video modeling. Additionally, we propose an optimization-driven sampling strategy that adaptively assigns denser sampling points to regions with higher temporal variation. Extensive experiments validate the effectiveness of our approach, demonstrating its superiority in video representation tasks. 

\noindent{\textbf{Limitation.}} The proposed optimization-driven sampling method dynamically inserts new nodes into high temporal variation segments. However, \textsc{node pruning} has not been incorporated into Tree-NeRV due to the absence of a well-defined pruning criterion, potential training instability, and the added complexity of hyperparameter tuning. Following the principle of `simple is better', we adopt a coarse-to-fine sampling strategy to balance efficiency and effectiveness. We encourage future research to further explore Tree-NeRV and develop a simple yet effective pruning method to enhance its performance.

{
    \small
    \bibliographystyle{ieeenat_fullname}
    \bibliography{main}
}

\setcounter{page}{1}
\maketitlesupplementary

\appendix
\section{Overview}
\label{sec:supp_overview}
In this supplementary document, we provide additional details to complement our main paper. First, we describe the lower and upper bound query mechanism in Tree-NeRV in~\cref{sec:supp_treenerv}. Next, we present an in-depth discussion and visualization of the balancing mechanism adopted in Tree-NeRV in~\cref{sec:supp_balance}. We then provide a preliminary review of NeRV in~\cref{sec:supp_nerv} and detail the implementation of our experiments in~\cref{sec:supp_setup}. Furthermore, we include additional experimental results to analyze our method in~\cref{sec:supp_additional result}. Finally, we provide qualitative comparisons in~\cref{sec:supp_Qualitative result}.

\section{Query Mechanism in Tree-NeRV}
\label{sec:supp_treenerv}
Tree-NeRV introduces a novel feature representation paradigm by organizing and storing features within a Binary Search Tree (BST), enabling efficient non-uniform sampling and retrieval. In a balanced Tree-NeRV, the query process follows a divide-and-conquer strategy, reducing the search space by half at each step. At each node, the query key is compared with the current node’s temporal key, and based on the result, the search proceeds to either the left or right subtree. Since an exact match is rare, the query typically continues until reaching a leaf node, which is the most common case in Tree-NeRV. Consequently, the query time complexity is determined by the height of the tree, yielding $\mathcal{O}(\log n)$ complexity. In contrast, \textsc{linked-list}-based representations, where features are sequentially stored, require a linear scan to locate the nearest temporal keys. This results in a worst-case query complexity of $\mathcal{O}(n)$, making retrieval significantly slower for large video sequences. 
To further optimize the search process, we introduce a set of intermediate variables that dynamically store and update the lower and upper bounds during traversal. This ensures that a single query efficiently retrieves both bounds. The query process summarized in~\cref{alg:search}.

\begin{algorithm}[t]
\caption{Temporal Embedding Query in Tree-NeRV}
\label{alg:search}
\begin{algorithmic}[1]
\State \textbf{Input:} Query time $t_i$, subtree $T$
\State \textbf{Output:} Time embedding $v_i$

\State Initialize traversal from the root node $(k, v)$ of $T$ 
\State Set $(k_i^l, v_i^l) \gets \text{None}, \quad (k_i^u, v_i^u) \gets \text{None}$  \Comment{Initialize bounds}

\While{$T \neq \emptyset$} 
    \If{$t_i = k$}  
        \State \Return $v_i = v$  \Comment{Exact match found}
    \ElsIf{$t_i < k$}  
        \State $(k_i^u, v_i^u) \gets (k,v)$  \Comment{Update upper bound}
        \State $T \gets T_L$  \Comment{Traverse left subtree}
    \Else  
        \State $(k_i^l, v_i^l) \gets (k,v)$  \Comment{Update lower bound}
        \State $T \gets T_R$  \Comment{Traverse right subtree}
    \EndIf
\EndWhile

\State $v_i \gets \textbf{Interpolate}(v_i^l, v_i^u, k_i^l, k_i^u, t_i)$  \Comment{Linear interpolation}
\State \Return $v_i$

\end{algorithmic}
\end{algorithm}

\section{Tree-NeRV's balance mechanism}
\label{sec:supp_balance}

To ensure Tree-NeRV remains balanced and supports efficient query operations, we implement an automatic rebalancing algorithm inspired by the self-balancing mechanism of AVL trees.
To formally define balance, we first define the height of each subtree $T$ denoted as $h(T)$, and the height of a subtree is the longest path from its root node to a leaf node, where a leaf node is defined as a node with both left and right subtrees empty, having a height of 0. The height of a subtree $T$ is computed recursively as:
\begin{equation}
    h(T) = 1 + \max\{h(T_L),  h(T_R)\}
\end{equation}

Beyond height, we define the balance factor $\beta$ for each subtree $T$ as the difference between the height of its left and right subtrees:
\begin{equation}
    \beta = h(T_L) - h(T_R)  
\end{equation}
To maintain Tree-NeRV’s balance, we enforce the AVL tree constraint, which requires each $\beta$ to satisfy:
\begin{equation} 
-1 \leq \beta \leq 1 
\end{equation}
any subtree $T$ with $\beta$ exceeds this range, is considered unbalanced, necessitating a rebalancing operation to restore its efficiency.

During training, Tree-NeRV encounters different types of imbalanced states. We categorize these scenarios and apply appropriate rebalancing strategies to restore balance efficiently, as summarized in~\cref{tab:imbalances}.

\begin{figure}
    \centering
    \includegraphics[width=0.5\textwidth]{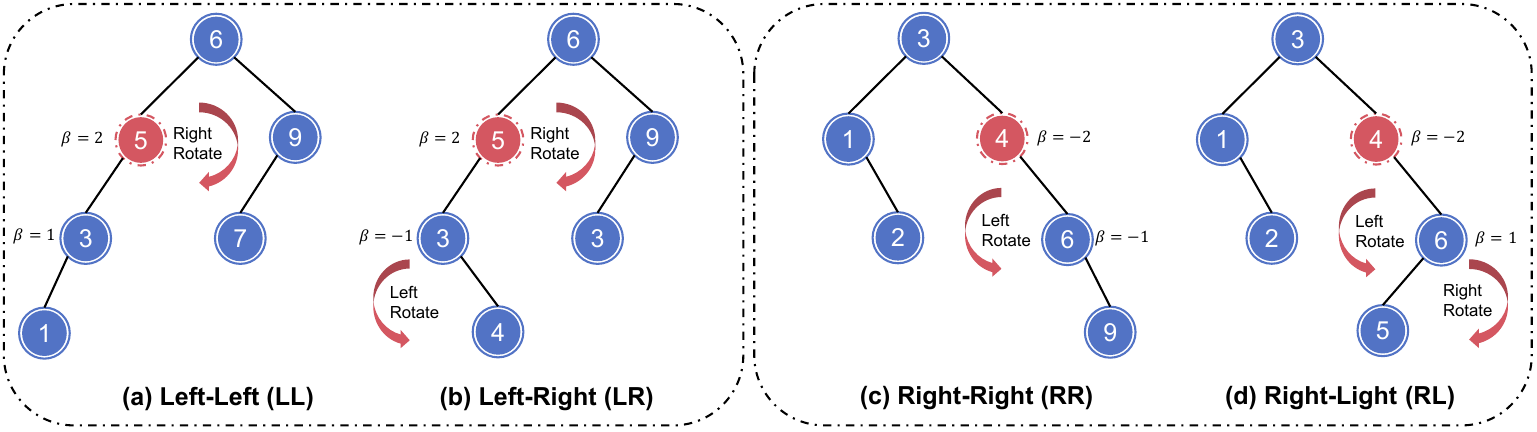}
\caption{\textbf{Right \& Left rotation adopted in Tree-NeRV.}
    }
  \label{fig:rot}
\end{figure}

\begin{table*}[t]
\centering
\caption{Classification of imbalance cases in Tree-NeRV and their corresponding rebalancing operations. Here, $ib$ denotes the imbalanced node, and $ibc$ denotes its child node.}
\label{tab:imbalances}
\begin{tabular}{ccc}
\toprule
\textbf{Imbalance Node} & \textbf{Child Node} & \textbf{Rebalancing Operation} \\ 
\midrule
$\beta_{ib} > 1$  & $\beta_{ibc} \geq 0$ & Right Rotation \\ 
$\beta_{ib} > 1$  & $\beta_{ibc} < 0$  & Left Rotation → Right Rotation \\ 
$\beta_{ib} < -1$ & $\beta_{ibc} \leq 0$ & Left Rotation \\ 
$\beta_{ib} < -1$ & $\beta_{ibc} > 0$  & Right Rotation → Left Rotation \\ 
\bottomrule
\end{tabular}
\end{table*}
To further illustrate these imbalance conditions,~\cref{fig:rot} provides a visual depiction of Tree-NeRV's rotation-based rebalancing operations. These scenarios are categorized based on the balance factor $\beta$, of the affected node and its child node. The corresponding rebalancing operations are applied as follows:
a) \textbf{Left-Left (LL)} Imbalance: The balance factor of node $5$ is $\beta_5=2$, indicating an imbalance. Its child node $3$ has $\beta_3=1$, leading to an LL imbalance. A single right rotation at node 5 restores balance.
b) \textbf{Left-Right (LR)} Imbalance: Similar to case (a), $\beta_5=2$, causing an imbalance. However, its child node $3$ now has $\beta_3=-1$, forming an LR imbalance. To restore balance, we first apply a left rotation at node $3$, followed by a right rotation at node $5$.
c) \textbf{Right-Right (RR)} Imbalance: The balance factor of node $4$ is $\beta_4=-2$, marking it as unbalanced. Its child node $6$ has $\beta_6=-1$, leading to an RR imbalance. A single left rotation at node $4$ restores balance.
(d) \textbf{Right-Left (RL)} Imbalance: Similar to case (c), $\beta_4=-2$, causing an imbalance. However, its child node $6$ now has $\beta_6=1$, forming an RL imbalance. To restore balance, we first apply a right rotation at node $6$, followed by a left rotation at node $4$.

\section{NeRV}
\label{sec:supp_nerv}

\begin{figure}[tbp]
  \centering
    \includegraphics[width=0.45\textwidth, height=0.18\textheight]{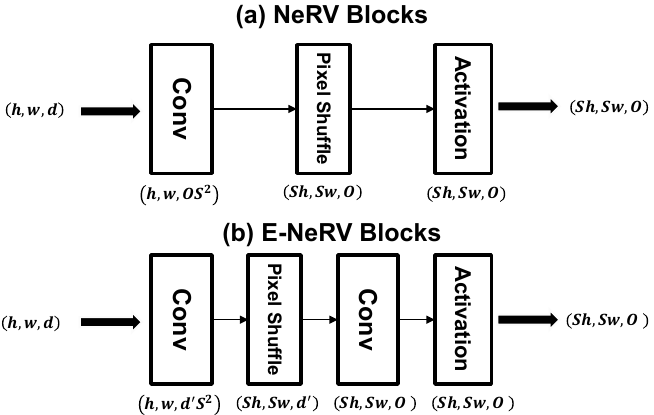}
    \caption{Comparision of NeRV and E-NeRV block.
    }
    \label{fig:supp_nerv}
\end{figure}
Neural Representations for Videos (NeRV)~\cite{chen2021nerv} is an Implicit Neural Representation (INR) framework designed for efficient video modeling. Unlike conventional INR methods that map spatiotemporal coordinates to pixel values, NeRV directly learns a frame-index-to-frame mapping, enabling fast and compact video representations.
Given an RGB video sequence $\mathbb{V}=\{x_t\}_{t=0}^{T-1}$, where each frame $x_t \in \mathbb{R}^{3 \times H \times W}$, NeRV formulates its mapping as: 
\begin{equation} 
x_t = f(\gamma(t)), 
\end{equation} 
where $f: \mathbb{R}^{t} \rightarrow \mathbb{R}^{3 \times H \times W}$ is the learnable function, and $\gamma(t)$ is an embedding function that encodes the frame index $t$ into a high-dimensional space.
The function $f$ is typically parameterized as a cascade of convolution-based NeRV blocks, which progressively upsample and refine feature representations. 

As illustrated in~\cref{fig:supp_nerv}, each NeRV block consists of: 
1) A convolutional layer to extract and transform features.
2) A pixel-shuffle upsampling operation~\cite{shi2016realtimesingleimagevideo} to increase spatial resolution.
3) An activation function (e.g., ReLU~\cite{glorot2011deep}, GELU~\cite{hendrycks2016gaussian}) to introduce non-linearity.
Given an input feature of shape $h \times w \times d$, NeRV aims to upsample it by a factor of $S$. The standard NeRV pipeline follows: 
\begin{equation} conv_{3 \times 3}(d, O S^2) \rightarrow \text{pixel-shuffle}(S). 
\end{equation} 

The number of trainable parameters in a single NeRV block is $3 \times 3 \times O \times d$.

To reduce parameter redundancy while maintaining performance, E-NeRV introduces an intermediate projection dimension $d'S^{2}$, modifying the block structure as:
\begin{equation} 
conv_{3 \times 3}(d, d'S^2) \rightarrow \text{pixel-shuffle}(S) \rightarrow conv_{3 \times 3}(d', O). 
\end{equation}
The total parameter count in an E-NeRV block is given by: $3 \times 3 \times d' S^2 \times (d \times S^2 + O)$. By selecting a smaller intermediate channel dimension $d'$, E-NeRV significantly reduces parameters while preserving spatial reconstruction quality.

While E-NeRV achieves substantial parameter savings, we observe that the expressiveness of NeRV blocks remains positively correlated with their parameter count. Excessive parameter reduction can degrade video reconstruction quality. To balance efficiency and performance, we adopt a hybrid design:
1) E-NeRV blocks are applied only in the first NeRV block, which requires the largest upsampling factor (e.g., $5\times$ scaling) and has the highest intermediate channel dimension.
2) tandard NeRV blocks are retained for all subsequent layers, preserving feature expressiveness while maintaining computational efficiency.

\section{Experimental Setup}
\label{sec:supp_setup}
\subsection{Implementation Details}

\noindent{\textbf{Baseline Implementation:}}
For HNeRV~\cite{chen2023hnerv}, and FFNeRV~\cite{lee2023ffnerv}, we conducted experiments using their publicly available implementations. FFNeRV adopts a multi-resolution feature grid, which we implemented following the original paper, using resolutions of [64, 128, 256, 512]. We controlled the parameter budget by adjusting the feature dimensions accordingly. 
For DS-NeRV~\cite{yan2024ds}, we developed our own implementation based on the open-source code of FFNeRV. Following the original work, we utilized varying numbers of static codes ($\sim \text{30–100}$) and dynamic codes ($\sim \text{150–400}$) to match their settings.
Notably, both HNeRV and DS-NeRV downscale video resolution to a fixed aspect ratio of 1:2 (i.e., height:width). However, in our experiments, we maintain the original 9:16 resolution for all video frames. To ensure a fair comparison, we adjusted their feature sizes accordingly, aligning with other methods in our evaluation.

\noindent{\textbf{Tree-NeRV Configuration:}}
In our implementation, we adjusted the number of channels in the latent features and NeRV blocks to control the model size, while keeping other hyperparameters consistent with the settings reported in the original papers.

For example, when processing a $1080 \times 1920 \times 3 \times 600$ UVG~\cite{mercat2020uvg} video sequence, we adopted a uniform sampling rate of $0.1$, resulting in $60$ initial features. These features were iteratively grown in four stages, with the top 10 GOPs being inserted with new nodes during each stage. Each feature code was represented as a $9 \times 16 \times 100$ three-dimensional vector. The NeRV blocks used stride steps of $5, 3, 2, 2, 2$, and the minimum number of channels was set to 72. Under this configuration, the total number of parameters amounted to approximately 3M.
For the DAVIS~\cite{wang2016mcl} dataset, which contains videos with fewer frames and higher dynamic variations, we adjusted only the number of initial features to $10$ and increased the feature dimension to $9 \times 16 \times 120$. All other settings were kept consistent with those used for the UVG dataset.

\begin{table}[!ht]
    \centering
    \tabcolsep=0.12cm
    \resizebox{1.0\linewidth}{!}{
        \begin{tabular}{c|ccccccc}
        \hline
            Video & size & resolution & $h_s\times w_s\times dim_s$ & init feature & topk & $Ch_{min}$ & strides  \\ \hline
            Beauty & 3 & $1080\times 1920$ & $9\times 16\times 100$ & 60 & 10 & 64 & (5,3,2,2,2)  \\ 
            Bosph & 3 & $1080\times 1920$ & $9\times 16\times 100$ & 60 & 10 & 64 & (5,3,2,2,2)  \\ 
            Honey & 3 & $1080\times 1920$ & $9\times 16\times 100$ & 60 & 10 & 64 & (5,3,2,2,2)  \\ 
            Yacht & 3 & $1080\times 1920$ & $9\times 16\times 100$ & 60 & 10 & 64 & (5,3,2,2,2)  \\ 
            Ready & 3 & $1080\times 1920$ & $9\times 16\times 100$ & 60 & 10 & 64 & (5,3,2,2,2)  \\ 
            Jockey & 3 & $1080\times 1920$ & $9\times 16\times 100$ & 60 & 10 & 64 & (5,3,2,2,2)  \\ 
            Shake & 3 & $1080\times 1920$ & $9\times 16\times 100$ & 30 &20 & 64 & (5,3,2,2,2)  \\ \hline
        \end{tabular}
    }
    \caption{Architecture details of Tree-NeRV on UVG.}
    \label{tab:uvgstucture}
\end{table}

\begin{table}[!ht]
    \centering
    \tabcolsep=0.12cm
    \resizebox{1.0\linewidth}{!}{
        \begin{tabular}{c|ccccccc}
        \hline
            Video & size & resolution & $h_s\times w_s\times dim_s$ & init feature & topk & $Ch_{min}$ & strides  \\ \hline
            Blackswan & 3 & $1080\times 1920$ & $9\times 16\times 120$ & 10 & 10 & 72 & (5,3,2,2,2)  \\ 
            Bmx-trees & 3 & $1080\times 1920$ & $9\times 16\times 120$ & 10 & 10 & 72 & (5,3,2,2,2)  \\ 
            Boat & 3 & $1080\times 1920$ & $9\times 16\times 120$ & 10 & 10 & 72 & (5,3,2,2,2)  \\ 
            Breakdance & 3 & $1080\times 1920$ & $9\times 16\times 120$ & 10 & 10 & 72 & (5,3,2,2,2)  \\ 
            Camel & 3 & $1080\times 1920$ & $9\times 16\times 120$ & 10 & 10 & 72 & (5,3,2,2,2)  \\ 
            Car-roundabout & 3 & $1080\times 1920$ & $9\times 16\times 120$ & 10 & 10 & 72 & (5,3,2,2,2)  \\ 
            Car-shadow & 3 & $1080\times 1920$ & $9\times 16\times 120$ & 10 &10 & 72 & (5,3,2,2,2)  \\ 
            Cows & 3 & $1080\times 1920$ & $9\times 16\times 120$ & 10 &10 & 72 & (5,3,2,2,2)  \\ 
            Dance & 3 & $1080\times 1920$ & $9\times 16\times 120$ & 10 &10 & 72 & (5,3,2,2,2)  \\ 
            Dog & 3 & $1080\times 1920$ & $9\times 16\times 120$ & 10 &10 & 72 & (5,3,2,2,2)  \\ \hline
        \end{tabular}
    }
    \caption{Architecture details of Tree-NeRV on Davis.}
    \label{tab:davisstructure}
\end{table}

\section{Additional Experiments}
\label{sec:supp_additional result}
\subsection{Topk Selection}
\begin{figure*}
    \centering
    \includegraphics[width=1\textwidth]{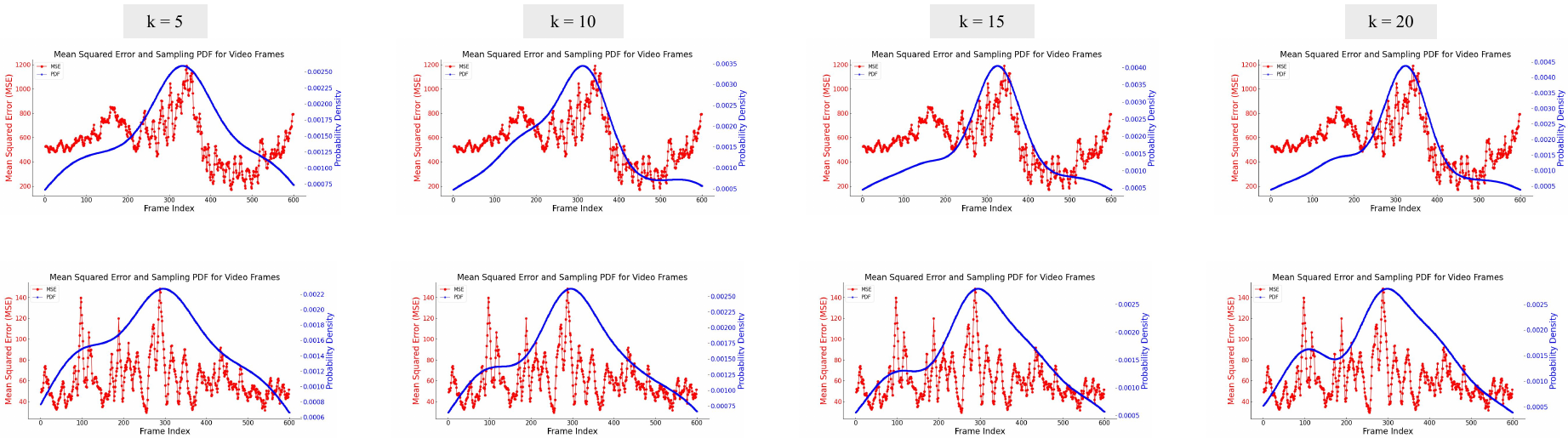}
\caption{\textbf{Tree-NeRV sampling results under different setting of \textbf{topk}, on Jockey (top), Beauty (bottom).}
    }
  \label{fig:supp_topk}
\end{figure*}

To further investigate Tree-NeRV, we conducted an additional ablation study. First, we evaluated the effect of different Top-$k$ values on Tree-NeRV's sampling behavior and compression performance. Specifically, we tested $k$ values of 5, 10, 15, and 20, analyzing Tree-NeRV's reconstruction results on the UVG dataset. As shown in \cref{tab:supp_abla}, similar compression performance was achieved across the different $k$ values. Additionally, in \cref{fig:supp_topk}, we visualized the actual sampling outcomes of Tree-NeRV for each $k$ setting, observing a consistent sampling trend across the different configurations.
\begin{table}[!t]
    \centering
    \tabcolsep=0.15cm
    \renewcommand{\arraystretch}{1.2} 
    \begin{tabular}{c|ccc|c}
        \hline
        Topk & 5 & 15 & 20 & 10\\ 
        \hline
        PSNR & 33.21 & 33.30 & 33.32 & \textbf{33.36}\\
        \hline
    \end{tabular}
    \caption{Ablation study for feature length on UVG. Ours sampled 160 feature points after training.}
    \label{tab:supp_abla}
    \vspace{-3mm}
\end{table}

\subsection{Video Compression}

Our compression pipeline follows a standard three-step process: global parameter pruning, quantization, and entropy-based encoding. \Cref{tab:supp_compression} presents the impact of these compression techniques on the final results. Moving forward, we aim to integrate more advanced compression strategies into NeRV-like approaches to further optimize efficiency. Additionally, we plan to explore node pruning as a mechanism for reducing video stream redundancy within Tree-NeRV.

\begin{table*}[t]\small
    \setlength{\tabcolsep}{3pt}
  \centering
  \begin{tabular}{l|ccccccc}
     UVG                 & Beauty & Bospho & Honey & Jockey & Ready & Shake &Yacht \\
    \midrule[1.5pt]
    N/A                           &33.54/0.920 &35.63/0.965 &39.88/0.990 &32.74/0.912 &26.86/0.861 &35.28/0.953 &29.74/0.908  \\
    8-bit Quant                   &33.48/0.919 &35.55/0.965 &39.86/0.989 &32.71/0.912 &26.79/0.860 &35.27/0.953 &29.68/0.908\\
    8-bit Quant + Pruning (10\%)  &33.13/0.916 &35.27/0.964 &39.15/0.989 &32.08/0.911 &26.42/0.859 &35.04/0.953 &29.44/0.908 \\
  \end{tabular}
  \caption{Compression ablations on UVG in PSNR/SSIM.}
  \label{tab:supp_compression}
\end{table*}

\subsection{Perceptional Quality Comparison}
In the field of compression, a widely recognized trade-off exists between `rate-distortion-realism'~\cite{blau2019rethinkinglossycompressionratedistortionperception}. Given that Tree-NeRV is fully trained using the Mean Squared Error (MSE) loss, we aim to evaluate its performance not only in terms of distortion but also in perceptual realism. To this end, we adopt the Learned Perceptual Image Patch Similarity (LPIPS)~\cite{zhang2018unreasonableeffectivenessdeepfeatures} metric to assess the perceptual quality of Tree-NeRV on the UVG and Davis dataset. The results are shown in~\cref{tab:supp_lpips}.
\begin{table*}[!t]
    \centering
    \resizebox{\linewidth}{!}{ 
    \begin{tabular}{l|cccccc|c}
        \hline
        Video & Bosph & Honey & Shake & b-dance & b-swan & c-shadow & Avg. \\ 
        \hline 
        HNerv~\cite{chen2023hnerv} & 0.335$\pm$0.009 & 0.199$\pm$0.004 &0.242$\pm$0.018 &0.228$\pm$0.007 &0.367$\pm$0.006 &0.334$\pm$0.012 &0.284$\pm$0.009 \\
        \hline 
        \textbf{Ours} &\textbf{0.283}$\pm$\textbf{0.012} & \textbf{0.194}$\pm$\textbf{0.005} &\textbf{0.241}$\pm$\textbf{0.018} &\textbf{0.168}$\pm$\textbf{0.006} &\textbf{0.291}$\pm$\textbf{0.008} & \textbf{0.263}$\pm$\textbf{0.012} &\textbf{0.240}$\pm$\textbf{0.01} \\
        \hline
    \end{tabular}
    }
    \captionsetup{font=scriptsize}
    \caption{Additional LPIPS ($\downarrow$) results with both method set to 3M parameters.}
    \label{tab:supp_lpips}
    \vspace{-3mm}
\end{table*}

\section{Additional Qualitative Results}
\label{sec:supp_Qualitative result}
\subsection{Visualization of Video Representation}
\begin{figure*}
    \centering
    \includegraphics[width=1\textwidth]{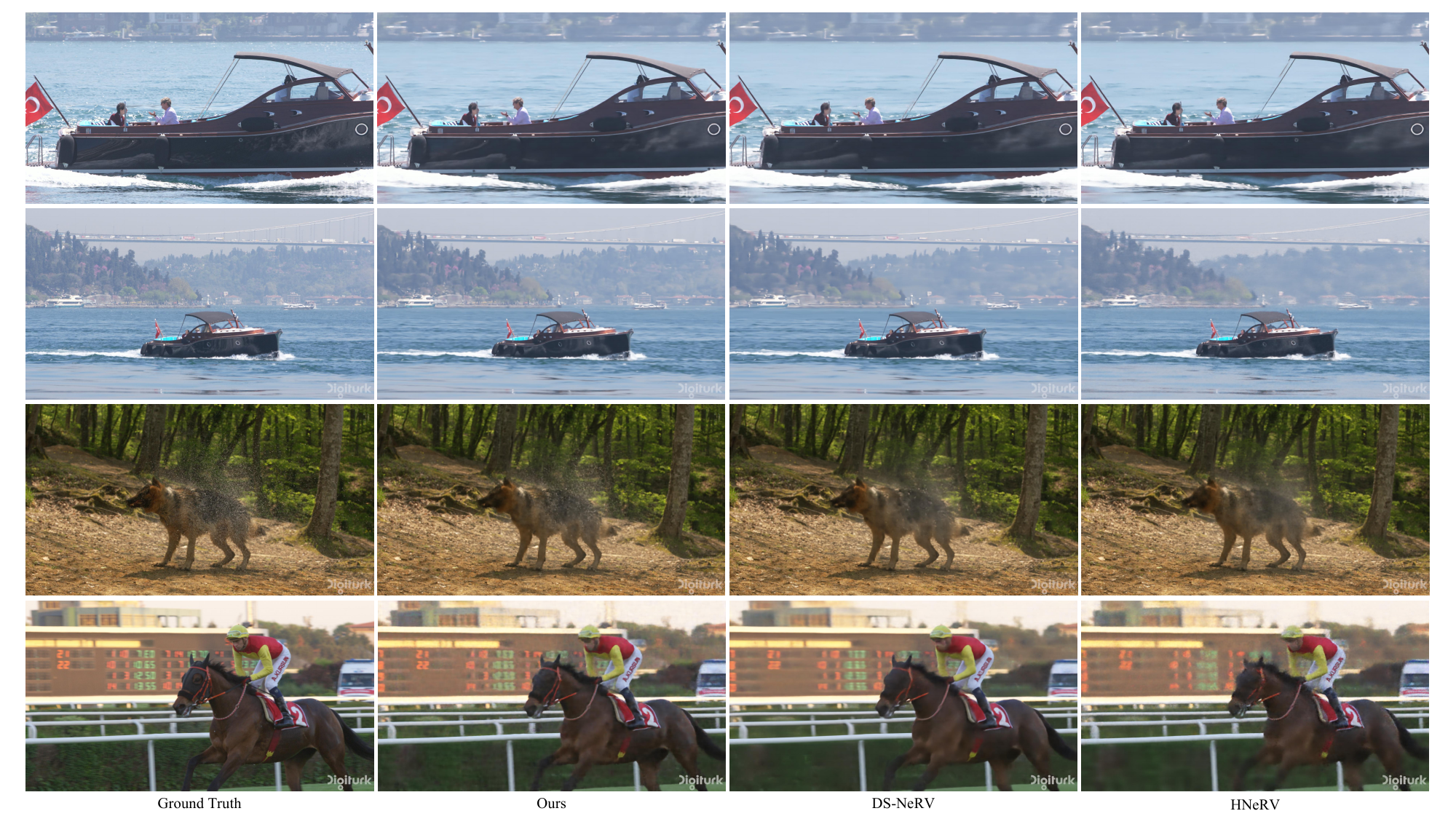}
\caption{\textbf{Additional video reconstruction results on UVG.}
    }
  \label{fig:supp_uvg}
\end{figure*}

\begin{figure*}
    \centering
    \includegraphics[width=1\textwidth]{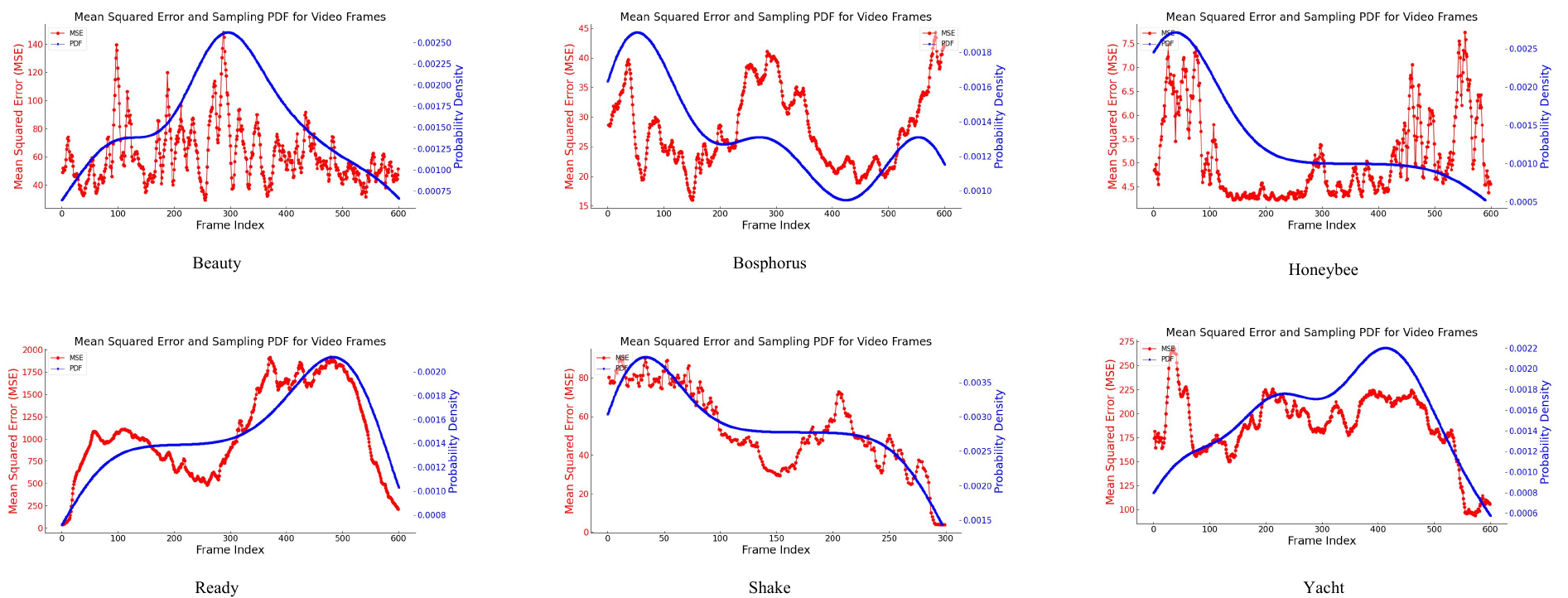}
\caption{\textbf{Additional Tree-NeRV sampling results on UVG.}
    }
  \label{fig:supp_sampling}
\end{figure*}

\begin{figure*}
    \centering
    \includegraphics[width=1\textwidth]{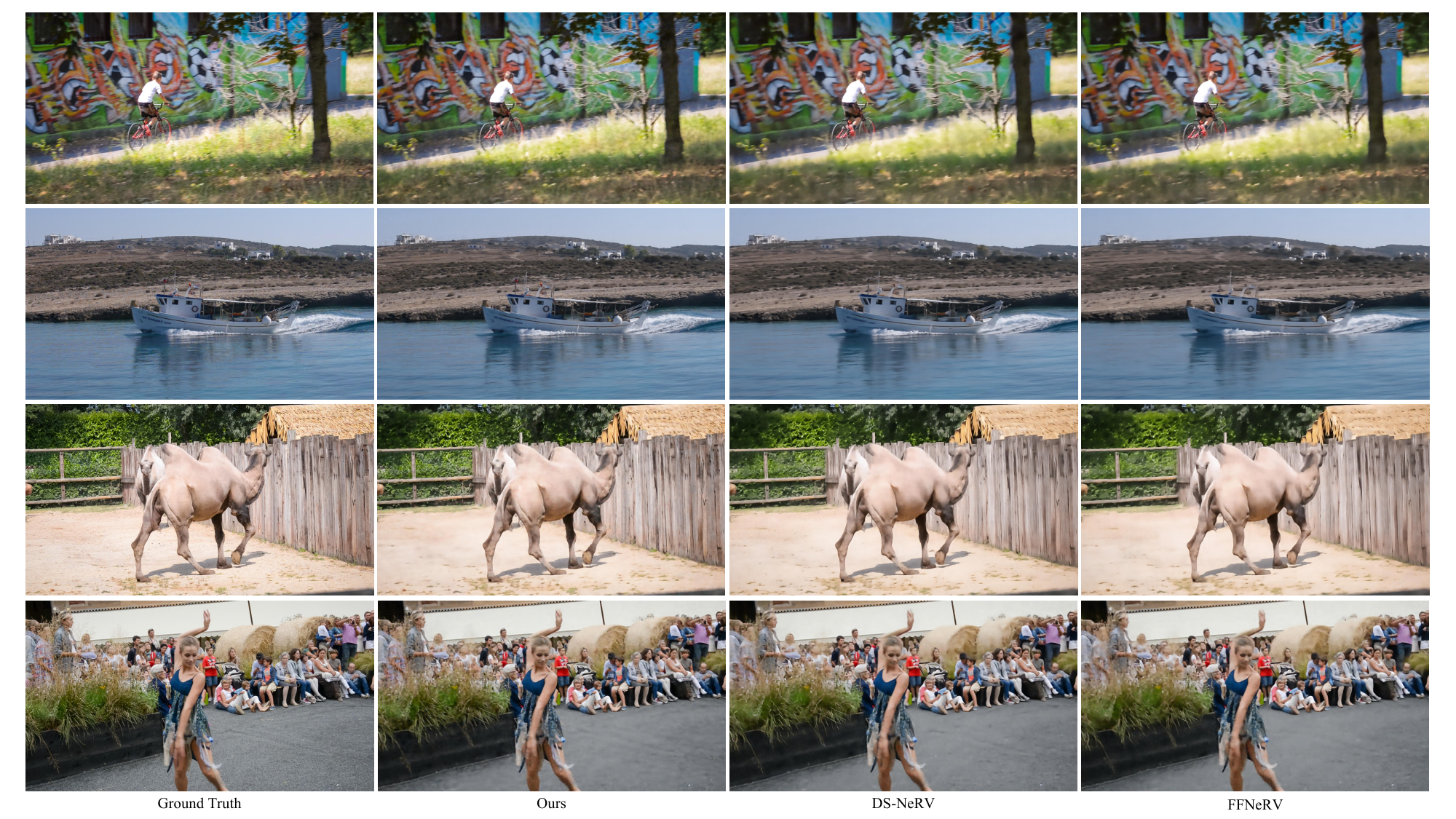}
\caption{\textbf{Additional video reconstruction results on DAVIS.}
    }
  \label{fig:supp_davis}
\end{figure*}

We present additional qualitative comparisons of video representation on the UVG and DAVIS datasets. Tree-NeRV consistently demonstrates superior reconstruction quality. For example, in \cref{fig:supp_uvg}, the first row highlights the circular rings on the boat, while the second row shows detailed high-frequency variations in the background. The third row captures splashing water, and the fourth row restores the numbers on the scoreboard. In \cref{fig:supp_davis}, Tree-NeRV outperforms other methods in reconstructing graffiti on the wall (first row), background architecture (second row), and the detailed textures on the camel (third row). As shown in~\cref{fig:supp_sampling}, these improvements are attributed to the tree-structured feature representation and our adaptive sampling strategy, which effectively captures the temporal redundancy in video streams.

\subsection{Visualization of Video Interpolation}
\begin{figure*}
    \centering
    \includegraphics[width=1\textwidth]{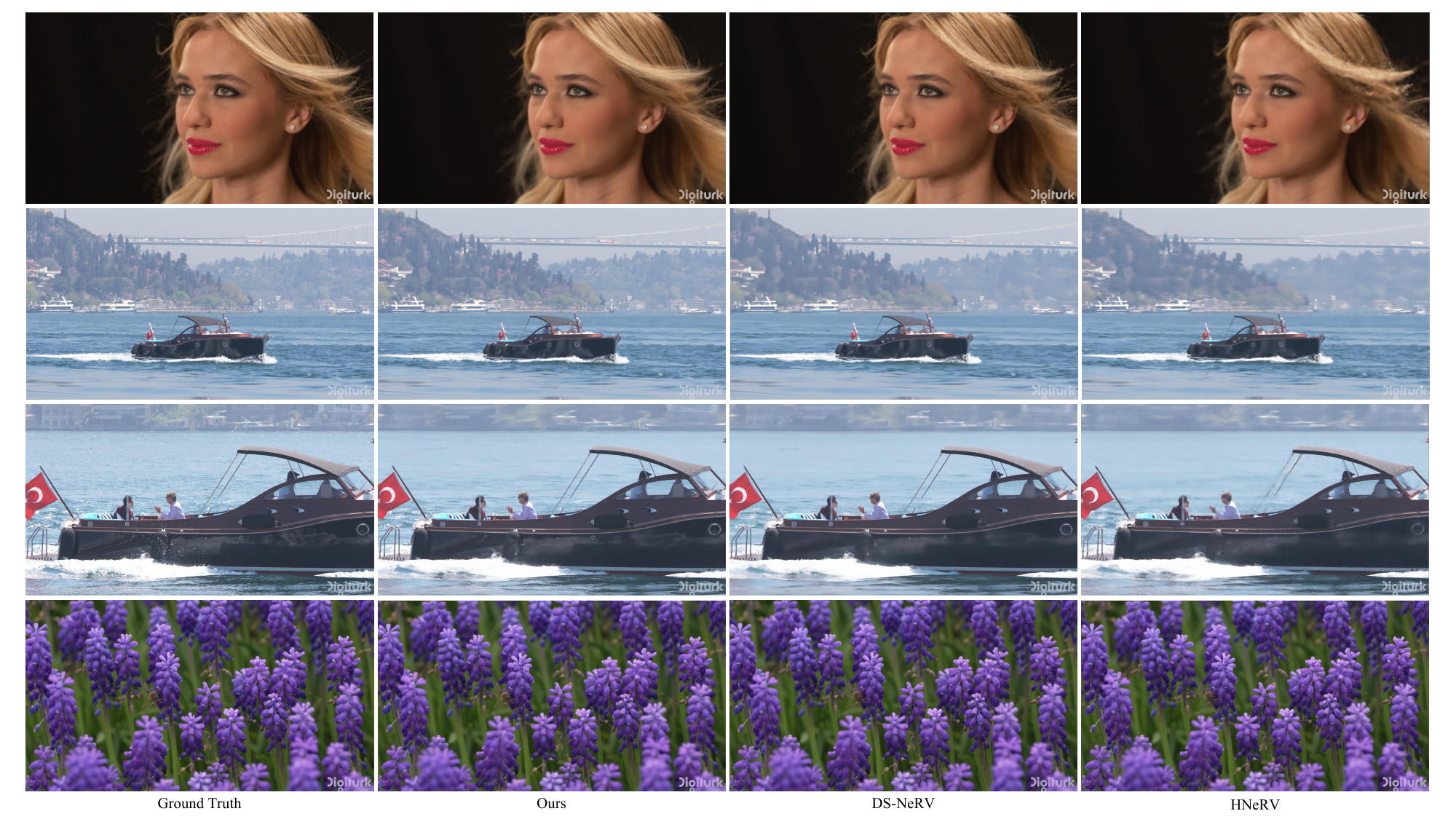}
\caption{\textbf{Additional video interpolation results on UVG.}
    }
  \label{fig:supp_interpo}
\end{figure*}

Additional visual comparisons of video interpolation results are available in \cref{fig:supp_interpo}. Tree-NeRV successfully preserves intricate details in previously unseen frames, demonstrating its superior interpolation capabilities.

\end{document}